\documentclass[11pt]{article}

\usepackage{acl}

\usepackage{times}
\usepackage{latexsym}

\usepackage[T1]{fontenc}

\usepackage[utf8]{inputenc}

\usepackage{microtype}

\usepackage{inconsolata}

\usepackage{graphicx}

\usepackage{amsmath}
\usepackage{amsthm}

\usepackage[normalem]{ulem}

\usepackage{xcolor} 
\usepackage{multirow}
\newtheorem{definition}{Definition}[section]
\newtheorem{assumption}[definition]{Assumption}
\newtheorem{lemma}[definition]{Lemma}

\usepackage{booktabs}

\newcommand{\xxx}[1]{\textcolor{black}{#1}}

\newcommand{\jkim}[1]{\textcolor{black}{#1}}

\newcommand{\rev}[1]{\textcolor{black}{#1}}
\newcommand{\seo}[1]{\textcolor{black}{#1}}

\title{Speculative Verification: \\
Exploiting Information Gain for Speculative Decoding}

\author{
  Sungkyun Kim \\
  Hanyang University \\
  \texttt{cheezestick@hanyang.ac.kr} \And
  Jaemin Kim \\
  Seoul National University \\
  \texttt{woals174@snu.ac.kr} \And
  Dogyung Yoon \\
  Hanyang University\\
  \texttt{sb0636@hanyang.ac.kr} \AND
  Jiho Shin\thanks{This work was done during his internship at the Machine Learning Systems Lab, Seoul National University.} \\
  KAIST\\
  \texttt{sjh010529@kaist.ac.kr} \And
  Junyeol Lee \\
  Hanyang University\\
  \texttt{shie007@hanyang.ac.kr} \And
  Jiwon Seo\thanks{Corresponding author.} \\
  Seoul National University \\
  \texttt{seojiwon@snu.ac.kr} \\
  }

\begin{document}
\maketitle
\begin{abstract}
Speculative decoding (SD) improves LLM inference latency by speculatively generating multiple tokens with a small draft model and verifying them with a larger target model. However, when speculation accuracy is low, the overhead from rejected tokens can negate its benefits, especially at large batch sizes.

We propose Speculative Verification (SV), an efficient augmentation to SD that predicts speculation accuracy and dynamically adapts the verification length to maximize throughput. SV introduces a small companion model, similar in size to draft model, to reduce uncertainty in speculation accuracy.
By exploiting the information gain from observing the companion distribution, SV reduces wasted verification on rejected tokens and improves decoding efficiency.

We evaluate SV across publicly available LLMs on seven NLP tasks using over a hundred combinations of draft, companion, and target models, including 13B--72B target models spanning base, instruction-tuned, and task-specific fine-tuned variants. Compared to target-only decoding, standard SD, and state-of-the-art SD variants, SV consistently delivers higher throughput across batch sizes. SV improves SD performance by up to 1.9$\times$, with an average 1.4$\times$ speedup at large batch sizes, showing robust and scalable gains for practical LLM inference.

\end{abstract}

\section{Introduction}
\label{sec:intro}

Large Language Models (LLMs) are widely used across many application domains, but their size and computational cost make large-scale inference serving a significant challenge. In particular, LLMs rely on autoregressive decoding -- generating one token at a time -- so producing $k$ tokens requires $k$ sequential steps, leading to GPU resource underutilization and increased latency.

Speculative Decoding\,(SD)\,\cite{leviathan2023fast} addresses this problem by using a small draft model to speculatively generate tokens, which are then verified in parallel by a larger target model. Because the draft model is fast and the target model validates multiple tokens in a resource-efficient manner, overall latency is reduced. However, if drafted tokens are rejected, both their verification and the recomputation incur additional overhead.

SD’s effectiveness depends on speculation accuracy, i.e., the fraction of drafted tokens accepted by the target model. Low accuracy negates its benefits; if most drafted tokens are rejected, or only a small fraction are rejected at large batch sizes (where SD’s gain already drops), the verification overhead can make SD slower than target decoding. Speculation accuracy depends on the alignment between the draft and target models, which fluctuates due to their capability gaps and input context variations. 

Identifying when these distributions align makes it possible to adjust the speculation length to minimize verification overhead for rejected tokens. However, predicting this alignment is challenging due to the complexity of LLM inference. Prior approaches attempt to predict acceptance using the draft model’s internal signals\,\cite{agrawal2024adaedl, zhang2025draft} or past acceptance history\,\cite{liu2024optimizing}, but our evaluation shows that these methods become ineffective \xxx{as batch size increases}.

In this paper, we propose speculative verification (SV), an approach to reliably predict speculation accuracy and maintain SD’s performance gains. Building on an information-theoretic foundation, SV compares the draft model’s output distribution with that of a similarly-sized {\it companion} model. By quantifying the alignment of their distributions, SV estimates the likelihood that the target model will accept the drafted tokens. Using these estimates, SV dynamically adjusts the verification length, minimizing verification cost for tokens likely to be rejected. This reduces the overhead of misaligned speculation and enables SD to scale effectively for real-world inference serving.

This paper contributes the concept of speculative verification (SV) and an optimized scheduling strategy and implementation for SV. 
Evaluations with publicly available LLMs across seven NLP tasks show up to 1.9$\times$ speedup over SD, and an average of 1.4$\times$ in large-batch settings (32–64).

The rest of the paper is structured as follows. Section\,\ref{sec:related} provides background information and related work on SD. Section\,\ref{sec:uncertainty} discusses uncertainty in speculation accuracy. Section\,\ref{sec:sv} presents our proposed SV technique. Section\,\ref{sec:scheduling} details our optimized scheduling approach and Section\,\ref{sec:impl} describes our prototype implementation. Section\,\ref{sec:eval} evaluates our methods, and Section\,\ref{sec:concl} concludes the paper.

\section{Background and Related Work}
\label{sec:related}

Speculative decoding (SD) reduces latency by using a small draft model to generate tokens speculatively, which are verified in parallel by the target model.
Because SD’s performance depends on speculation accuracy, prior work has focused on improving this accuracy or predicting it to adapt the speculation length; we review these works below. 

\noindent
{\bf Distillation and Self-Speculative Models.} To improve draft-target similarity, prior work has applied knowledge distillation, as in DistillSpec\,\cite{zhou2024distillspec} and HRSS\,\cite{zhang2025learning}. Because distillation requires costly fine-tuning, self-speculative models have been proposed to perform speculation mostly using the target model itself, thereby reducing additional training overhead.

LayerSkip\,\cite{layerskip} and Kangaroo\,\cite{liu2024kangaroo} use the first few layers of the target model for drafting. Medusa\,\cite{cai2024medusa} employs the full target model with additional decoding heads to predict multiple \jkim{draft} tokens in parallel. Eagle-3\,\cite{li2025eagle} further uses hidden states from multiple target-model layers in its decoding heads and generates draft token trees instead of linear sequences to better utilize hardware resources, especially for single-batch inference.

\noindent
{\bf Predicting Speculation Accuracy.} To predict token acceptance probability, SmartSpec\,\cite{liu2024optimizing} uses a moving average of recent acceptance rates and adjusts the draft length accordingly. Tetris\,\cite{wu2025tetris} and DySpec\,\cite{xiong2025dyspec} predict token acceptance using the draft model's probability for sampled tokens and prioritize verification accordingly. However, their approach relies on the strong assumption that the draft model’s token probability is well correlated with acceptance probability, which often does not hold in practice. 

To address this limitation, SVIP\,\cite{zhang2025draft}, AdaEDL\,\cite{agrawal2024adaedl}, and DISCO\,\cite{mamou2024dynamic} propose predicting acceptance based on the entropy of the draft model’s token distribution. These methods are motivated by the fact that token acceptance can be bounded using the KL divergence between the draft and target distributions. However, this bound is known to be loose\,\cite{canonne2022short, zhang2025draft}, and in practice the target distribution is unavailable during drafting. Hence, these methods estimate the bound using only the draft distribution’s entropy, leading to inaccurate acceptance prediction, as we show in Section\,\ref{subsec:eval-sdvariants}.

Staged speculation introduces mid-sized models to verify drafted tokens before final verification by the target model \cite{spector2023accelerating, syu2025hierarchical}. Rather than predicting speculation accuracy, the intermediate model performs speculative sampling on the draft output to reduce the target’s verification overhead. However, performance is fundamentally limited by the capability of the intermediate model and can even be worse than target-only decoding, as shown in our evaluation.

\begin{figure}[h]
  \includegraphics[width=0.9\linewidth]{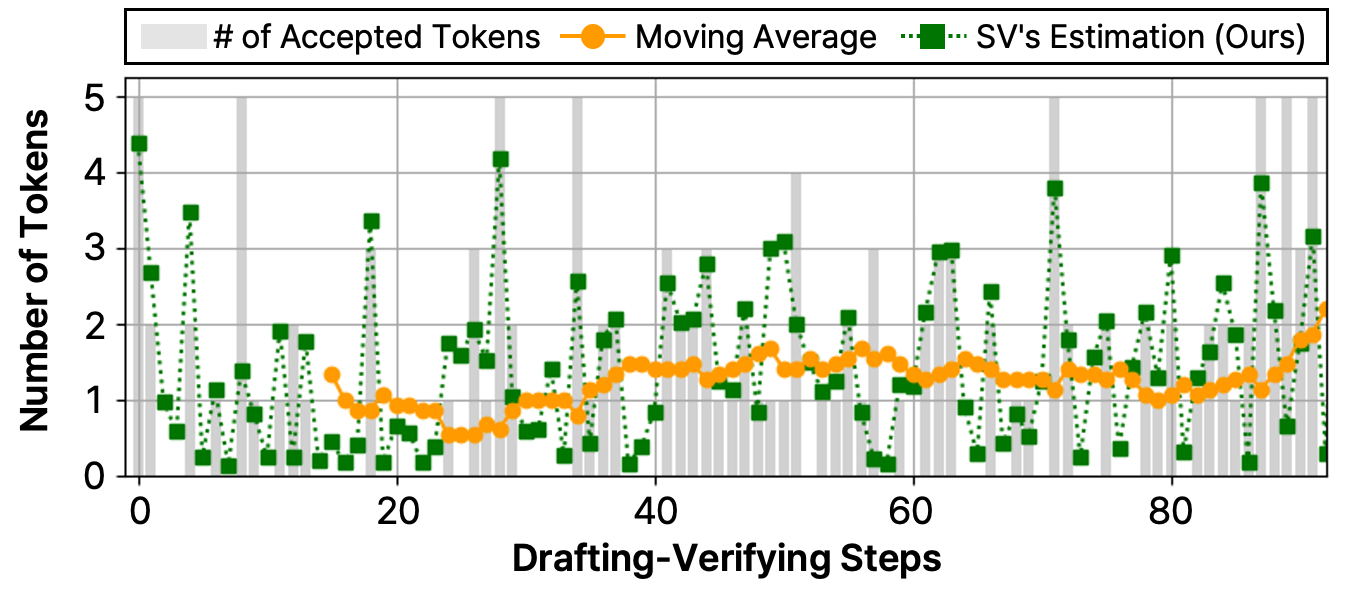}
  \caption{Accepted tokens per SD step:\,actual~vs.~estim.}
  \label{fig:uncertainty-expr}
\end{figure}

\section{Uncertainty in Speculation Accuracy}
\label{sec:uncertainty}

Since draft models are less capable than target models, their speculation accuracy is often inconsistent and uncertain\,\cite{liu2024optimizing, chen2025spin}. Predicting speculation accuracy could improve SD's effectiveness, so we explored whether it can be inferred using only the draft model's information. To investigate this, we ran SD on 128 prompts across two NLP tasks and analyzed the resulting speculation accuracy. Figure\,\ref{fig:uncertainty-expr} shows a subset of the results for one representative query over 100 draft-verification steps (draft length=5). The gray bars (accepted tokens per step) indicate that speculation accuracy fluctuates sharply and unpredictably across steps.

To understand what causes the accuracy changes, we examined tokens before and after sharp changes in accuracy. However, these tokens do not share any commonality in their embeddings or semantics (based on human interpretation). Moreover, we found that majority of accuracy-changing positions involve high-frequency tokens, such as stop words, which suggests that the observed fluctuations are not likely driven by meaning or context. A prior study proposed predicting speculation accuracy from recent history using a moving average\,\cite{liu2024optimizing}, but as shown in Figure\,\ref{fig:uncertainty-expr}, this estimation deviates from the actual accuracy. We also tested entropy-based predictions (e.g. SVIP), which showed limited accuracy\,(see Section\,\ref{subsec:eval-sdvariants}).

This uncertainty leads to a significant waste of verification cost. In our preliminary experiment, we found that over 40\% of verification effort was spent on rejected tokens, and that 48\% of SD steps were more expensive than decoding directly with the target model. We also observed that at larger batch sizes, SD’s already reduced performance gains are further offset by the cost of running the draft model and the additional verification overhead, which can result in overall performance degradation.

Predicting speculation accuracy is difficult due to the inherent complexity of token generation in LLMs. For instance, attention heads across layers serve different roles that vary with context\,\cite{clark2019does, wang2024differentiation}. Some compute attention scores globally over many tokens, while others focus locally on a subset. A single attention head can switch unpredictably between global and local computations\,\cite{donhauser2025unveiling}. 
Consistent with prior findings, our analysis correlating attention heads between the draft and target models shows that draft-side attention patterns do not provide reliable signals for predicting speculation accuracy.

From these preliminary analyses, we found that predicting speculation accuracy based on the draft model’s token distribution (or its entropy), attention patterns, or past accuracy is infeasible. This uncertainty in speculation accuracy poses a major challenge to scaling inference serving with SD.

\begin{figure*}[t]
  \centering
  \includegraphics[width=0.85\textwidth]{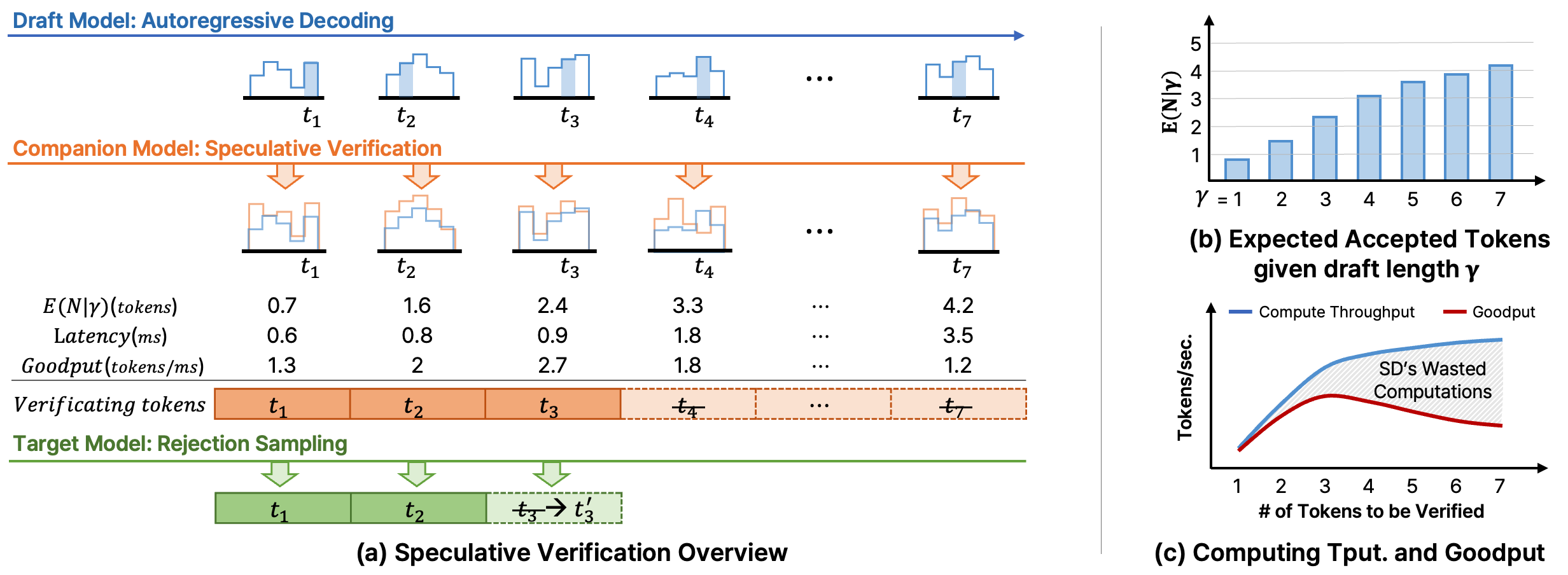}
  \caption{Example of selecting the verification length to maximize goodput (accepted tokens/sec) in SV.}
  \label{fig:overview}
\end{figure*}

\section{Introducing Speculative Verification}
\label{sec:sv}

To address speculation uncertainty in SD, we propose using additional information, as the draft model alone cannot reliably predict speculation accuracy. We extract this information from another LLM instance of similar size to the draft model. By comparing their token distributions, we aim to reduce uncertainty in token acceptance while preserving the target model’s output distribution.

 \subsection{Information Gain for Efficient Speculation}

SD accelerates inference because the draft model’s token distribution is reasonably aligned with that of the target model. However, this alignment is inconsistent, leading to unpredictable speculation accuracy, as discussed in Section\,\ref{sec:uncertainty}. If the accuracy could be predicted, the speculation length could be adjusted dynamically to minimize wasted verification cost and maximize SD efficiency.



To predict speculation accuracy, we introduce a small auxiliary model, namely a {\it companion} model, similar in size to the draft model. We assume that comparing the token distributions of the draft and companion models provides positive information for predicting token acceptance by the target model. \rev{We emphasize that this requirement is strictly weaker than requiring high correlation between the draft–companion and draft–target distribution similarities; SV only requires that observing the companion model's distribution provides \emph{positive} information gain for predicting target acceptance.} This assumption holds as long as the two models are not statistically independent.
Since modern LLMs are typically trained on partially shared datasets (e.g., Wikipedia and C4), statistical independence is unlikely in practice.
We further demonstrate in our evaluation that, across all examined combinations of draft, companion, and target models (trained by the same or different entities), SV yields consistently positive information gain\rev{, even when the correlation between draft--companion and draft--target similarity is low (see appendix\,\ref{appendix:infogain-vs-correlation} for details)}.


More formally, we exploit the information gain from knowing the distribution similarity between the draft and companion models to reduce the uncertainty in speculation accuracy. If a random variable $X$ represents the speculation accuracy, i.e., the acceptance probability of a token generated by the draft model, and $Y$ denotes the corresponding distribution for the token in the companion model, then the uncertainty of $X$ is measured as the entropy $H(X)$, and the conditional uncertainty is $H(X|Y)$ representing the remaining uncertainty of $X$ given the value of $Y$.

We aim to maximize the information gain $I(X;Y) = H(X) - H(X|Y)$, i.e., the amount of uncertainty reduced in speculation accuracy when knowing the companion model's distribution. We carefully choose $Y$ -- what to observe in the companion model's distribution (details discussed in the next section) -- so that we can accurately predict the acceptance probability of drafted tokens.

With this prediction, we adjust the verification length (see Section\,\ref{sec:scheduling}) to maximize SD efficiency. Because the companion model helps determine which tokens to verify or discard, we call this method speculative verification (SV).
Note that SV differs significantly from staged SD, where an intermediate model performs speculative sampling on the draft output. Their approach assumes that the intermediate model is better aligned with the target model, but its efficiency is limited by the intermediate model's capability, as shown in our evaluation.

\subsection{Indicators in Companion Model}
\label{subsec:indicator}

What should we observe and define as the conditioning random variable to minimize speculation uncertainty? The requirements are: (1) it must be strongly correlated with the acceptance probability of the drafted tokens, and (2) it must be simple to measure and quantify. We propose observing the joint condition of two variables: one ($S$) measures the distributional similarity between the draft and companion models, and the other ($A$) measures the draft token's acceptance probability under speculative sampling in the companion model (although this sampling is not actually performed). More formally, \xxx{${\small S}$ is a natural divergence in SD}, i.e., ${\small S = \sum_{i \in \text{vocab}} \min\left(P_{\text{d}}(t_i), P_{\text{c}}(t_i)\right)}$ and ${\small A = \min\left(1, \frac{P_{\text{c}}(t_{\text{d}})}{P_{\text{d}}(t_{\text{d}})}\right)}$, where $t_{\text{d}}$ is a drafted token, and $P_{\text{d}}$ and $P_{\text{c}}$ are the token distributions of the draft and companion models, respectively.

$S$ denotes the similarity of the draft and companion token distributions at the current decoding step. When these distributions are similar -- i.e., the draft and companion models agree on likely next tokens -- the target model's distribution is also likely similar, as the current context is relatively easy to decode even for less-capable models. Thus, $S$ serves as an indicator of draft–target distribution similarity.


\seo{We validated the effectiveness of the indicator $S$ by measuring how much uncertainty it reduces in a token's acceptance probability. As detailed in Appendix~\ref{appendix:correlation}, knowing $S$ yields an information gain of \rev{4--13\%}. We also compared the similarity between the draft and companion token distributions with that between the draft and target distributions. Although SV does not rely on a strong correlation between these similarities, we still observed strong correlations of 0.75--0.87 across three draft--companion--target model groups.}




However, distribution similarity alone is not sufficient to minimize speculation uncertainty, as the specific drafted token also has a significant impact on its acceptance probability. To account for this, we incorporate the drafted token's acceptance probability under the companion model as an additional conditioning variable $A$. Together, this probability and the distribution similarity $S$ yield a reliable predictor of token acceptance, substantially reducing speculation uncertainty.

We validated experimentally that these two variables significantly reduce speculation uncertainty. The details are in Section\,\ref{subsec:eval-infogain}, but using $S$ and $A$ reduces speculation uncertainty by 34\% and improves the target model's acceptance rate by 20\%.

\seo{Because SV only decides which drafted tokens are sent to the target model for verification, it does not alter the sampling distribution. If SV forwards a drafted token for verification, the token is accepted or rejected according to the standard SD's sampling rule, preserving the target model’s distribution. If SV drops a drafted token, the target model generates the corresponding position directly from its own distribution. Therefore, in all cases, the resulting token follows the target model’s distribution.}


\subsection{\rev{Companion Model Acquisition}} 
\seo{For a given pair of draft and target models, a companion model can be obtained by fine-tuning or quantizing the draft model, or by selecting a publicly available model of similar size. We evaluate these options using publicly available models and report the corresponding information gain in Appendix~\ref{appendix:info-gain-subsection}. To briefly summarize the results, when the companion model is fine-tuned from the draft model, the corresponding information gain is 0.26 on average. When the companion model is obtained by quantizing the draft model, the information gain is 0.4 for the case where the draft model is Qwen2.5-1.5B, the companion model is its 4-bit quantized version, and the target model is Qwen2.5-14B. For companion models that are neither fine-tuned nor quantized from the draft model, the average information gain is also 0.26, which is as good as the overall average information gain across all evaluated model pairs.}

\section{Scheduling for Speculative Verification}
\label{sec:scheduling}



We now explain how to take advantage of the predicted acceptance probability to maximize effective throughput in terms of accepted tokens. We refer to this variant of throughput as goodput in this paper -- the number of accepted tokens per unit time. To achieve high goodput, we must determine the optimal subset of drafted tokens to verify in the target model, i.e., optimal verification length.

We need to consider two factors to optimize verification length:
1) wasted computation on rejected tokens, and 2) GPU's resource utilization for a given verification length.  
Our goal is to balance these factors -- minimizing wasted computation while maximizing GPU resource utilization -- to achieve optimal goodput. We do this by estimating the expected number of accepted tokens for each possible verification length and selecting the length that maximizes estimated goodput (i.e., the expected number of accepted tokens divided by verification latency) as shown in Figure\,\ref{fig:overview}(b)\,and\,2(c).

We first explain how to compute the expected number of accepted tokens for a given verification length. Let $T_i$ denote the random variable for the $i$'th token in the draft model, $P(T_i | S, A)$ denote its conditional acceptance probability in the target model (given information gain from the companion model), $N$ denote the number of accepted tokens in the target model, and $\gamma$ denote the number of drafted tokens verified in the target model.

\vspace{3pt}
The probability for $N$ given $\gamma$ is calculated as: \\
\vspace{-1pt}
{\small
\[
  \label{eq:prob_gamma}
  P_{\gamma}(N) =
    \begin{cases}
    P(T_{N+1}\!\neq\!t_{N+1}) \prod_{i=1}^{N}P(T_i\!=\!t_i)  & \text{if } N < \gamma,\\
    \prod_{i=1}^{\gamma} P(T_i\!=\!t_i)                               & \text{if } N = \gamma
    \end{cases}
\]
}where we use ${\small P(T_i|S,A)}$ instead of ${\small P(T_i)}$ for a more accurate acceptance probability, or its empirical estimate $\hat{P}(T_i|S,A)$, which is obtained from sampled data.
Then, the expected number of accepted tokens conditioned on $\gamma$ is ${\small E(N|\gamma) = \sum_{i=1}^{\gamma} i \cdot P_{\gamma}(N=i)}$.

Using the expected acceptance length $E(N|\gamma)$, or its estimate $\hat{E}(N|\gamma)$, we calculate goodput based on the profiled latency for a given verification length $\gamma$. We then vary $\gamma$, compute the corresponding goodput, and select the length that maximizes it. Figure\,\ref{fig:overview} illustrates this process of computing token acceptance probability, expected acceptance length, and the optimal $\gamma$ for maximum goodput.

We find the optimal $\gamma$ by incrementally increasing it while goodput improves; once goodput declines, we revert to the previous $\gamma$ as the optimal value. This approach works because goodput is concave with respect to the verification length. As we increase the verification length from a small value, latency grows slowly at first, since the GPU’s compute resources are not fully utilized. Once the verification length is large enough for full resource utilization, latency increases proportionally, but the expected acceptance length grows more slowly -- the cumulative probability of accepted tokens diminishes as more drafted tokens are included.

\paragraph{Scheduling for Batch Execution.}

To extend our approach to batch-level optimization, we apply the same goodput-based strategy used for single queries. We use a greedy algorithm that starts with an empty verification token sequences and iteratively adds the token that yields the highest increase in expected acceptance length, regardless of which query it belongs to. This process continues as long as goodput improves and stops once it reaches its peak. While this strategy may prioritize some queries over others, it does not cause starvation. In practice, token acceptance rates drop sharply beyond a certain length, and verification always yields at least one accepted token per query, ensuring forward progress (see the fairness study in our evaluation).

\section{Implementation}
\label{sec:impl}

We implemented standard LLM inference optimizations in our vLLM-based system, including input tensor compaction (to avoid padding for variable-length verification), data-parallel drafting (reusing multiple GPUs allocated for tensor-parallel verification for drafting), and CUDA graph execution (to reduce kernel launch overhead; applied to all baselines -- SD, SV, and other SD variants). In doing so, we also fixed a FlashInfer bug that miscomputed attention scores when CUDA graph capture was used with padded inputs\,\cite{ye2025flashinfer}.

We further optimize performance by overlapping the target model's verification with the drafting and speculative verification of the next iteration. To enable this, we run the target and draft/companion models in separate processes using NVIDIA’s Multi-Process Service (MPS), which allows them to share GPU resources. We configure MPS so that the draft/companion uses a small fraction (30\%) of the resources, while the target uses the remainder -- or optionally all resources. 
Since this optimization involves running two micro-batches concurrently, we monitor the memory overhead of maintaining their contexts and adjust batch sizes accordingly. To accurately profile verification overhead, we measure it while the draft/companion are running to capture interference effects, which remain consistent as their workload sizes (i.e., the number of tokens processed by the draft and companion models) are constant.

\begin{figure*}[t]
  \includegraphics[width=0.95\textwidth]{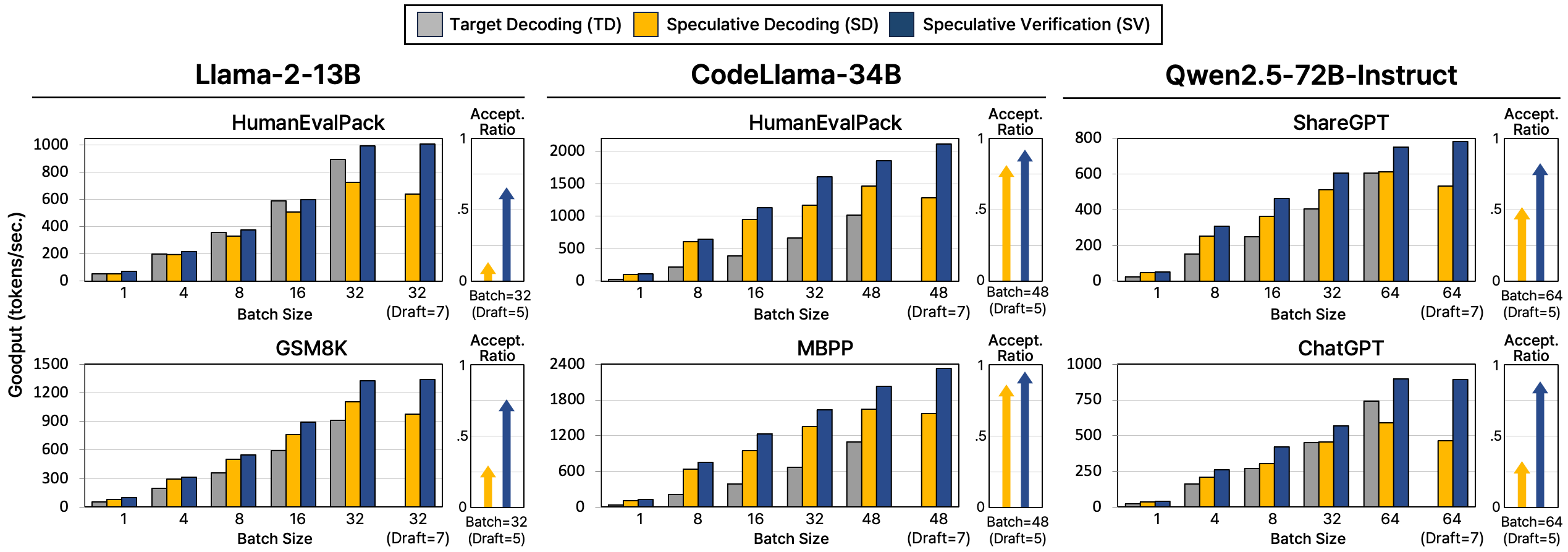}
  \caption {Goodput {\small (accepted tokens/sec)} of target-only decoding, SD, and SV across three target models and six tasks.}
  \label{fig:overall}
\end{figure*}

\section{Evaluation}
\label{sec:eval}

\subsection{Evaluation Settings}

We evaluate SV against target-only decoding, standard SD, and five SD variants (including SVIP and Eagle-3). Our evaluation uses two widely adopted LLM families, Qwen (v2.5, v3) and Llama (v2, v3), with 104 draft/companion/target combinations and five target sizes (13B -- 72B), covering base, instruction-tuned, and task-tuned models across seven tasks. Detailed settings are in the appendix.



\begin{figure}[t]
  \centering
  \includegraphics[width=\linewidth]{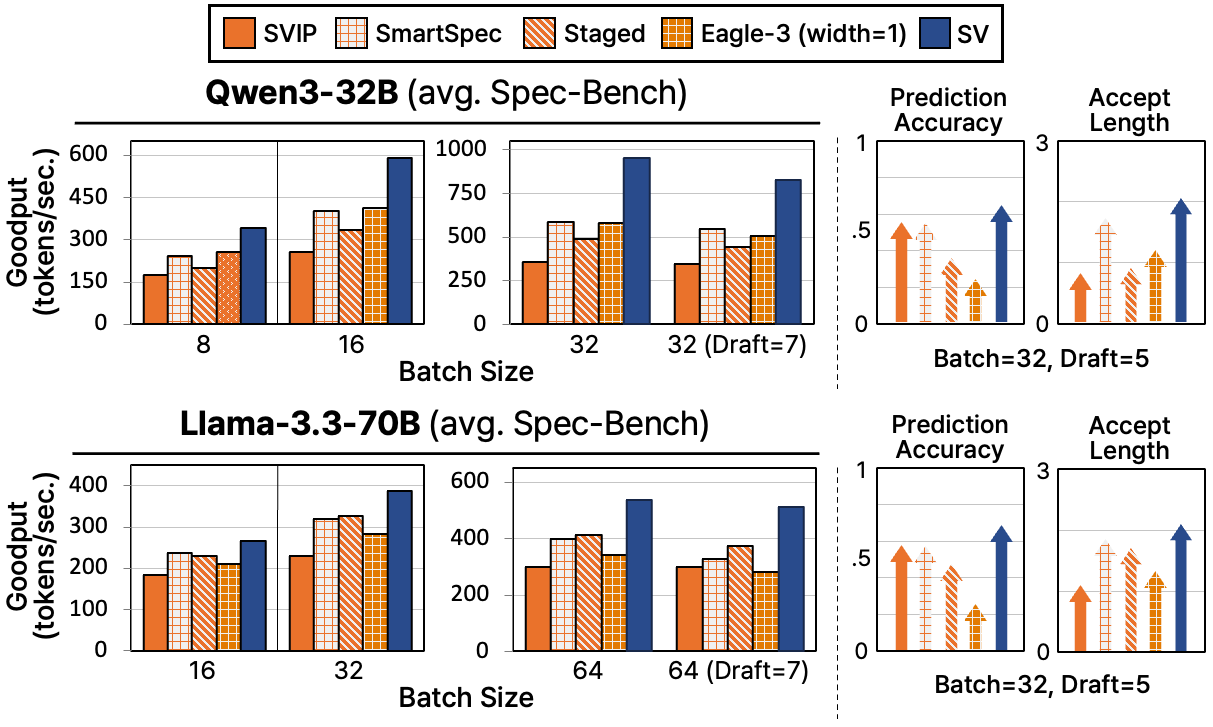}
  \caption{Goodput of SOTA SD variants and SV.}
  \label{fig:sdvariants}
\end{figure}

\subsection{Overall Performance Evaluation}
\label{subsec:eval-overall}

We evaluated the token generation throughput of SV, comparing it to target-only decoding and SD. Using draft lengths of 5 and 7 tokens, we increased the batch size from 1, doubling up to 32--64, the maximum supported by the GPU’s memory.

Figure\,\ref{fig:overall} shows results for a representative subset of models and tasks (other results omitted due to space limits) with draft length of 5. In all experiments, SV consistently outperforms both SD and target decoding. At the maximum supported batch sizes, SV is on average 1.4$\times$ faster than SD, with peak speedups reaching 1.9$\times$; at the same batch sizes, SV also achieves up to 4.5$\times$ higher token accept rate (indicated by the arrow-shaped bars). As batch size increases, SD’s performance gains decrease and can even fall below that of target decoding -- in such cases, SV outperforms SD by a large margin. Moreover, for difficult tasks, such as GSM8K and ChatGPT, where SD is known to perform poorly\,\cite{syu2025hierarchical, chen2025spin}, SV continues to deliver strong performance.

\subsection{Comparison to SOTA SD Variants}
\label{subsec:eval-sdvariants}
We compare SV with other state-of-the-art (SOTA) SD variants, including SVIP, SmartSpec, Staged SD, and Eagle-3. 
We use the same previous evaluation settings. For Eagle-3, we use the authors' official models and set the tree width to 1 for batch sizes$>$1 following the authors' setting. For the other variants, whose implementations are not publicly available, we implemented them following the descriptions in their papers. We verified that our implementations match the reported single-batch performance; multi-batch performance is not reported for most prior methods.

Figure\,\ref{fig:sdvariants} shows results for two target models on Spec-Bench, with full per-subtask results reported in the appendix. Across all batch sizes and models, SV consistently outperforms the other SD variants. Compared to the best SD variant, SV is up to 1.61$\times$ faster for Qwen{\small(batch=32, draft=5)} and 1.37$\times$ faster for Llama{\small(batch=64, draft=7)}. At the maximum batch sizes, SV is 1.3--2.64$\times$ faster than the SD variants.

At large batch sizes, the efficiency of these variants degrades for different reasons. For SVIP and SmartSpec, draft lengths can vary widely across queries in the same batch, causing a small number of long drafts to incur large drafting overhead.
For Staged SD, performance is limited by the capability of the intermediate model, which becomes significant as batch size grows. For Eagle-3, speculation accuracy is low; while token-tree generation mitigates this issue for single-batch inference, its overhead becomes dominant at larger batch sizes.

\seo{Because the companion model incurs the highest relative overhead for prefill-heavy, short-output workloads, we report separate evaluation results for these cases. Specifically, for the two Spec-Bench subtasks with these characteristics -- RAG (average input length: 703 tokens; average output length: 53 tokens) and Summarization (average input length: 712 tokens; average output length: 121 tokens) -- SV still achieves an average speedup of 1.39$\times$ over SD at the largest evaluated batch size, as shown in Figure~\ref{fig:appendix_performance} in Appendix~\ref{appendix:additional-eval}.}

\subsection{Information Gain from $S$ and $A$}
\label{subsec:eval-infogain}

In this section, we quantify the information gain in the random variable $X$, i.e., the acceptance probability of a drafted token in the target model, by comparing the token distributions of the draft and companion models. We observe two variables -- $S$ and $A$: $S$ measures the distributional similarity between the two models, and $A$ is the draft token’s acceptance probability based on the companion model, assuming SD’s sampling rule is applied.

To measure the uncertainty of $X$ (i.e., entropy) and the information gain from observing $S$ and $A$, we perform inference using the Llama2 13B/160M/68M models, applying SD on the ShareGPT and HumanEvalPack datasets.
Specifically, we generate tokens using SD’s process with two draft lengths (5 and 7). For the drafted tokens, we observe $S$ and $A$ using the companion model, but do not apply SV’s optimization of verification lengths. We then use the target model to measure the acceptance probability of the drafted tokens based on SD’s acceptance rule. After evaluating acceptance, we repeat the SD's standard drafting and verification steps.

\begin{table}[t]
    \small
    \centering
    \setlength\tabcolsep{2pt}
    \begin{tabular}{crlrlrl}
    \hline
       & \multicolumn{2}{c}{\textbf{Uncertainty}}     & \multicolumn{2}{c}{\textbf{Cond. Uncertainty}}    & \multicolumn{2}{c}{\textbf{Info. Gain}}   \\
      
      \textbf{Resolution} & \multicolumn{2}{c}{$H(X)$} & \multicolumn{2}{c}{$H(X|S,A)$} & \multicolumn{2}{c}{$I(X;S,A)$}   \\
    \cline{2-7}
     & \phantom{0}chat & code & \phantom{0123}chat & code & chat & code \\
    \hline
     5$\times$5 & 1.38 & 1.78 & 1.01 & 1.50 & 0.38 & 0.28 \\
    10$\times$10 & 1.38 & 1.78 & 0.97 & 1.48 & 0.41 & 0.31 \\
    20$\times$20 & 1.38 & 1.78 & 0.95 & 1.46 & 0.43 & 0.32 \\
    272 & 1.38 & 1.78 & 0.91 & 1.42 & 0.48 & 0.37 \\
    \hline
    \end{tabular}
    \caption{Uncertainty in the token acceptance probability and the information gain from observing $S$ and $A$.}
    \label{tab:entropy-comparison}
\end{table}

\begin{table}[t]
    \small
    \centering
    \begin{tabular}{lcc}
    \hline
    \textbf{Target Model} & \textbf{\# of D-C pairs} & \textbf{Info. Gain} \\
    \hline
    Llama-2-13B-chat        & 30 & 0.07 -- 0.59 \\
    Qwen2.5-14B-Instruct    & 30 & 0.03 -- 0.60 \\
    Llama-3.3-70B-Instruct  & 30 & 0.03 -- 0.40 \\
    \hline
    \end{tabular}
    \caption{Information gain across 90 draft-companion-target (D-C-T) models (full table in appendix\,\ref{appendix:info-gain-subsection}).}
    \label{tab:info-gain-all}
\end{table}

Table\,\ref{tab:entropy-comparison} shows the uncertainty of $X$, the conditional uncertainties when observing $S$ and $A$, and the information gain from observing the two variables. Since $S$ and $A$ are continuous, we discretize their ranges into equal-sized sub-ranges and report the corresponding conditional uncertainties at varying resolution levels (indicated in the leftmost column). We also include results using our adaptive binning, which assigns smaller bins where data samples are denser (the bottom row).
With adaptive binning, observing $S$ and $A$ gives an information gain of 30--40\% of the total entropy, indicating a strong relationship between these variables and $X$\,\cite{panzeri1996analytical, quinlan1996improved}.

Furthermore, we measured information gain more extensively across 90 publicly available draft–companion–target (D–C–T) model combinations using the Spec-Bench dataset. Table\,\ref{tab:info-gain-all} summarizes the observed ranges of information gain across these combinations, which cover nearly all publicly available configurations for the three target models (Llama-2-13B, Qwen2.5-14B, and Llama-3.3-70B). Across all 90 combinations, we confirmed that SV's information gain is consistently positive (see appendix\,\ref{appendix:info-gain-subsection} for details).

\begin{figure}[t]
  \centering
  \includegraphics[width=\linewidth]{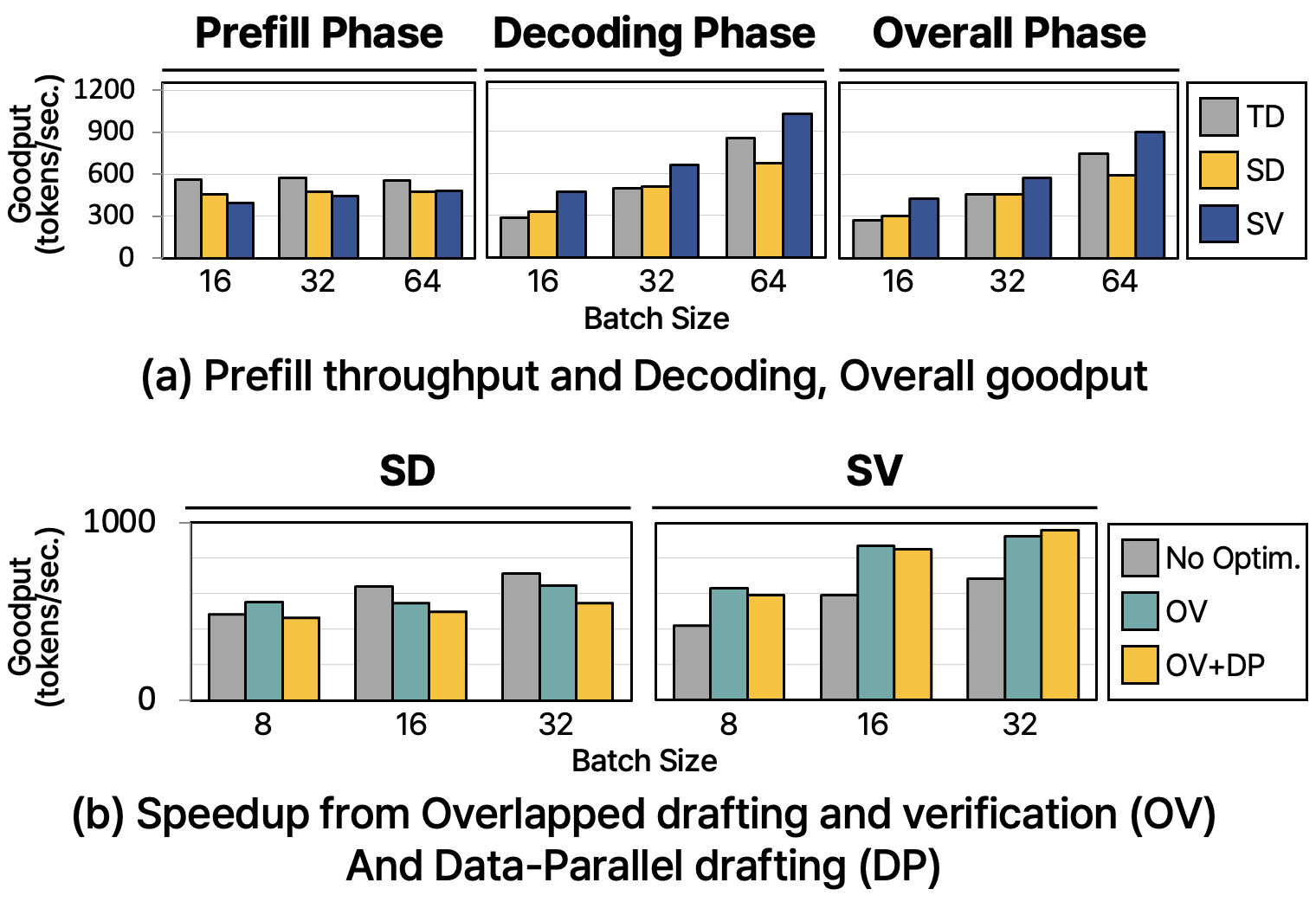}
  \caption{Performance Breakdown: Prefill vs. Decoding (top) and Runtime Optimizations (below)}
  \label{fig:breakdown}
\end{figure}

\subsection{Performance Breakdown}
\label{subsec:breakdown}

\noindent
{\bf Prefill and Decoding Performance.}
We separately measured prefill throughput (processed tokens/sec) and decoding goodput (generated tokens/sec) for SD and SV. Figure\,\ref{fig:breakdown} shows results for Qwen 72B/1.5B/0.5B models. 
Due to the companion model's overhead, SV's prefill throughput is lower than SD’s. We partially mitigate this by skipping logit computations in the draft and companion models during prefill, which reduces overhead by 3--5\%. Still, SV’s prefill throughput remains about 10\% lower than SD’s (30\% lower than target decoding). However, during the decoding phases, SV outperforms both SD and target decoding, especially at larger batch sizes. At a batch size of 32, SV is \xxx{25\%} faster than SD, and at 64, it is \xxx{52\%} faster. \xxx{Overall, SV achieves higher throughput consistently across all batch sizes.}

\noindent
{\bf Runtime Optimization Breakdown.} We also evaluated two major runtime optimizations: overlapping decoding with verification (OV) and data-parallel decoding (DP). We applied both optimizations to SD and SV and measured decoding goodput using Qwen 32B/1.5B/0.5B. 
Figure\,\ref{fig:breakdown}(b) compares the goodput of SD and SV. For SD, applying OV degrades performance due to the high cost of verification; that is, interference between verification and drafting reduces overall throughput.

With SV, applying OV consistently improves performance, as SV reduces the verification cost and thereby mitigates interference between drafting and verification. Applying DP on top of OV provides additional gains only at large batch sizes. At smaller batch sizes, however, the synchronization overhead of the companion model outweighs the benefits of parallelism.

\noindent
{\bf Reduction in Verification Cost.}
Table\,\ref{tab:companion-overhead} reports the reduction in verification cost achieved by SV. For Qwen2.5, CodeLlama, and Llama2, SV reduces verification cost by 21--41\% in terms of total computation, including both the prefill and decoding phases. In comparison, the companion model adds only 1.3--5.3\% to the total amount of computation, and most of this overhead is masked by overlapping its execution with verification, as discussed in Section\,\ref{sec:impl}. The memory overhead of the companion model ranges from 2.8\% to 8.1\% across the evaluated configurations.





\begin{table}[h]
  \centering
  \footnotesize
  \setlength\tabcolsep{2pt}
  \renewcommand{\arraystretch}{1.1}
  
  \setlength{\aboverulesep}{0pt}
  \setlength{\belowrulesep}{0pt}
  \setlength{\cmidrulekern}{0.4em} 

  \begin{tabular}{lcccccc}
    \hline
    \multirow{2}{*}{\textbf{Model (D/C/T)}} & \multicolumn{2}{c}{\textbf{Accept Rate}} & \multicolumn{3}{c}{\textbf{TFLOPs}} & \textbf{Reduction} \\
    \cmidrule(r){2-3} \cmidrule(l){4-6}
     & \textbf{SD} & \textbf{SV} & \textbf{SD} & \textbf{V} & \textbf{C}  & \textbf{in TFLOPs}\\
    \hline
    Qwen2.5 &  \multirow{2}{*}{0.56} & \multirow{2}{*}{0.62} & \multirow{2}{*}{78} & \multirow{2}{*}{16$\downarrow$} & \multirow{2}{*}{1$\uparrow$} & \multirow{2}{*}{$-$18.27\%} \\
    (1.5B/0.5B/32B) & & & & &  \\
    Qwen2.5 &  \multirow{2}{*}{0.47}& \multirow{2}{*}{0.57} & \multirow{2}{*}{84} & \multirow{2}{*}{21$\downarrow$} & \multirow{2}{*}{1$\uparrow$} & \multirow{2}{*}{$-$23.37\%} \\
    (0.5B/0.5B/32B) & & & & &  \\
    Qwen2.5 & \multirow{2}{*}{0.47} & \multirow{2}{*}{0.72}  & \multirow{2}{*}{84} & \multirow{2}{*}{19$\downarrow$} & \multirow{2}{*}{4$\uparrow$}  & \multirow{2}{*}{$-$17.43\%} \\
    (0.5B/1.5B/32B) & & & & &  \\
    CodeLlama       & \multirow{2}{*}{0.73} & \multirow{2}{*}{0.86} & \multirow{2}{*}{63}   & \multirow{2}{*}{20$\downarrow$} & \multirow{2}{*}{2$\uparrow$}  & \multirow{2}{*}{$-$27.63\%} \\
    (135M/1.2B/34B) & & & & &  \\
    Llama2          & \multirow{2}{*}{0.14} & \multirow{2}{*}{0.61} & \multirow{2}{*}{109}   & \multirow{2}{*}{50$\downarrow$} & \multirow{2}{*}{1$\uparrow$}  & \multirow{2}{*}{$-$44.63\%} \\
    (68M/160M/13B)  & & & & &  \\
    \hline
  \end{tabular}
  \caption{\seo{Comparison of the amount of computation in SD and SV, showing the reduction in verification cost ($\downarrow$) and the overhead of the companion model ($\uparrow$). Here, \(V\) denotes the reduction in verification cost, and \(C\) denotes the companion-model overhead. The values are averaged over Qwen on ShareGPT and Llama2 on CodeLlama with a draft length of 5.}}
  \label{tab:companion-overhead}
\end{table}

\subsection{Fairness in Verification Token Selection}
\label{subsec:fairness}

SV selects a subset of drafted tokens for verification and discards the rest to maximize goodput. Thus, some queries in a batch may have few or no tokens verified. While the target model's verification guarantees progress by generating at least one token per query (preventing starvation), we still evaluate the fairness of SV's verification token selection.
For this analysis, we ran generation with 1024 inputs and calculated the average number of tokens verified for each sequence. We then examined the five queries with the smallest average verification lengths. Compared to the overall average of 4.1 tokens, these bottom five queries had an average of 2.9 tokens verified; 39\% of their steps involved verifying 4--5 tokens, while 47\% involved 1--2 tokens, suggesting that their token allocation is fairly distributed. The full results are shown in the appendix, but SV's token selection is reasonably balanced and does not result in substantial unfairness.

\begin{figure}[t]
  \centering
  \includegraphics[width=\linewidth]{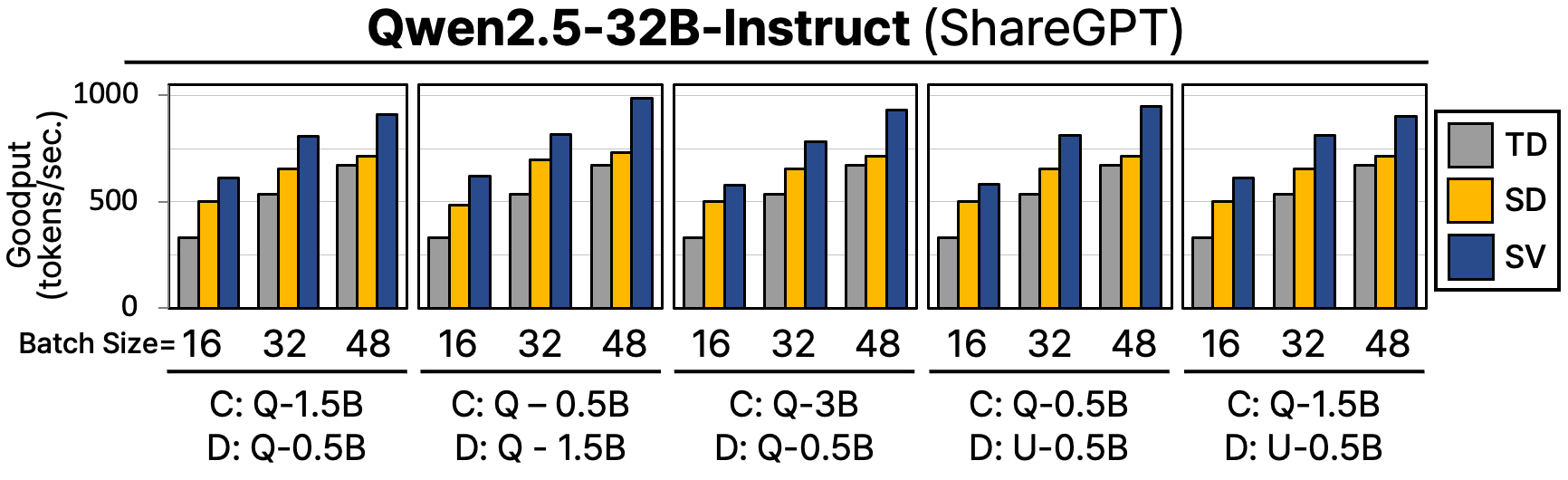}
  \caption{
  SV speedup across five draft-companion pairs of different model sizes and training entities {(Q:Qwen2.5, U:Unsloth; see appendix\,\ref{appendix:models} for details)}.}
  \label{fig:modelselection}
\end{figure}

\subsection{Robustness and Generality of SV}
\label{subsec:robustness}

\noindent
{\bf Effect of Draft/Companion Selection.} Using Qwen2.5-32B as the target model, we evaluated SV with five draft/companion pairs spanning different model sizes and training entities (0.5B, 1.5B, and 3B models trained by Alibaba and Unsloth; see appendix\,\ref{appendix:models}).  Figure\,\ref{fig:modelselection} shows that 
SV’s performance is robust to draft/companion choice, remaining consistent across all evaluated combinations.

\begin{figure}[t]
  \centering
  \includegraphics[width=\linewidth]{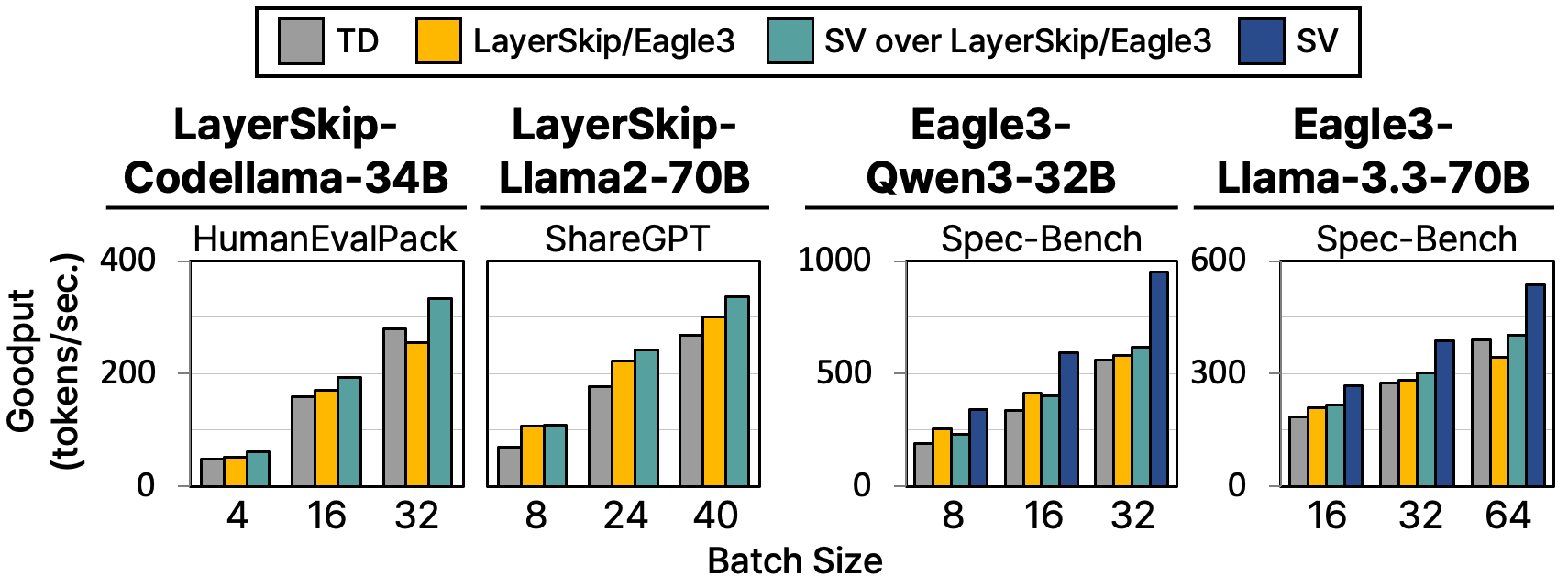}
  \caption{Performance of SV over LayerSkip/Eagle3.}
  \label{fig:selfspec}
\end{figure}

\noindent
{\bf SV for Self-Speculation.} We applied SV to two self-speculative models, LayerSkip and Eagle-3. For LayerSkip, we evaluated CodeLlama-34B and Llama2-70B (the drafting and companion layers specified in the appendix). For Eagle-3, we evaluated Qwen3-32B and Llama3.3-70B, using Qwen3-0.6B and Llama-3.2-1B as companion models.

Figure~\ref{fig:selfspec} shows the performance impact of applying SV to these models. For LayerSkip, SV yields substantial and consistent improvements, increasing performance at maximum batch sizes by \xxx{30\% for CodeLlama and 12\% for Llama-2.} 
\xxx{For Eagle-3, applying SV yields more modest gains and may degrade performance slightly at small batch sizes.}
This is due to the limited capability of Eagle-3's draft model, which achieves low acceptance rates of only 26\% and 27\% for drafted tokens. Using a stronger draft model would likely improve the synergy with SV. When we replaced the draft with a separate, more capable model, performance increased by up to 56\%, as indicated by the SV results in Figure~\ref{fig:selfspec}.

\section{Conclusion}
\label{sec:concl}

We presented Speculative Verification (SV), a technique that improves speculative decoding (SD) by adapting verification lengths based on predicted token acceptance. SV introduces a lightweight companion model and uses the alignment between its token distribution and that of the draft model to estimate draft–target alignment, which enables reliable prediction of speculation accuracy.

SV uses an information-theoretic framework to guide verification decisions, reducing wasted computation on rejected tokens and improving decoding efficiency, particularly at large batch sizes. Extensive evaluation across public LLMs, multiple NLP tasks, diverse draft–companion–target combinations, and batch sizes up to 64 shows that SV consistently outperforms target decoding, standard SD, and state-of-the-art SD variants, achieving up to 1.9$\times$ speedup over SD with an average 1.4$\times$ gain in high-throughput settings.

\section{Limitations}
\label{sec:limitation}

\noindent
{\bf Public Model Coverage.} Our evaluation is conducted using publicly available models, which may introduce model-specific biases. However, within this scope, we perform extensive evaluation: we report goodput improvements across 14 draft–companion–target model pairs from the HuggingFace model hub and also analyze speculation uncertainty reduction across more than 90 model pairs.

\noindent
{\bf Limited Variance Analysis.} We do not repeat the full generation workload multiple times per evaluation setting to report variance or confidence intervals. Each evaluation run processes multiple requests over approximately 10,000 decoding iterations, and we report the resulting average goodput. While we do not measure variance across repeated runs, the reported results are aggregated over a large number of decoding steps within each run.

\section*{Acknowledgments}

This work was supported by the New Faculty Startup Fund from Seoul National University and Institute of Information \& communications Technology Planning \& Evaluation (IITP) grant funded by the Korea government (MSIT) (No.RS-2025-02214497, 
No.RS-2025-02263167, 
IITP-2025-II211817 (ITRC), 
No.RS-2024-00438729, 
RS-2021-II211343). 
This work was also supported by Basic Science Research Program through the National Research Foundation of Korea(NRF) funded by the Ministry of Education (RS-2026-25476387), the research fund of Hanyang University (HY201700000002388), and Automation and System Research Institute at Seoul National University (No.0418-20250030). Jiwon Seo is the corresponding author.

\bibliography{custom}

\appendix

\begin{table*}[t]
  \centering
  \small
  \setlength\tabcolsep{1pt}
  \renewcommand{\arraystretch}{1.1}
  \footnotesize

\begin{tabular}{cccccc}
    \hline
    \textbf{Section} & \textbf{Dataset} & \textbf{GPU} & \textbf{Target Model} & \textbf{Companion Model} & \textbf{Draft Model} \\
    \hline
    
    \multirow{6}{*}{Sec.\,\ref{subsec:eval-overall}} & ChatGPT   & A100      & Qwen/                & Qwen/                 & Qwen/   \\
    & ShareGPT  &  $\times$4 & Qwen2.5-72B-Instruct & Qwen2.5-1.5B-Instruct & Qwen2.5-0.5B-Instruct \\ 
    \cline{2-6}
                            
     & HumanEvalPack & A100      & meta-llama/      & TinyLlama/                  & amd/   \\
    & MBPP          & $\times$2 & CodeLlama-34b-hf & TinyLlama\_v1.1\_math\_code & AMD-Llama-135m-code \\
    \cline{2-6}
    & HumanEvalPack     & A100      & meta-llama/      & JackFram/  & JackFram/    \\
    & GSM8K             & $\times$2 & Llama-2-13b-hf   & llama-160m & llama-68m \\
    
    \hline

    \multirow{8}{*}{\shortstack[c]{Sec.\,\ref{subsec:eval-sdvariants}\\ Sec.\,\ref{subsec:robustness}}} & \multirow{8}{*}{Spec-Bench} & \multirow{4}{*}{\shortstack[c]{A100\\$\times$4}} & \multirow{4}{*}{\shortstack[c]{meta-llama/\\Llama-3.3-70B-Instruct}} & meta-llama/ & meta-llama/ \\
    & & &  & Llama-3.2-3B-Instruct & Llama-3.2-1B-Instruct  \\
    \cline{5-6}
    & & & & meta-llama/ & yuhuili/ \\
    & & & & Llama-3.2-1B-Instruct & EAGLE3-LLaMA3.3-Instruct-70B  \\
    \cline{3-6}
    & & \multirow{4}{*}{\shortstack[c]{A100\\$\times$2}} & \multirow{4}{*}{\shortstack[c]{Qwen/ \\ Qwen3-32B}} & Qwen/ & Qwen/ \\
    & & & & Qwen3-1.7B & Qwen3-0.6B  \\
    \cline{5-6}
    & & & & Qwen/ & RedHatAI/ \\
    & & & & Qwen3-0.6B & Qwen3-32B-speculator.eagle3  \\

    \hline

    \multirow{4}{*}{Sec.\,\ref{subsec:robustness}} & \multirow{2}{*}{ShareGPT} & A40 & facebook & facebook & facebook \\
    & & $\times$4 & layerskip-llama2-70B & layerskip-llama2-70B(5)$^{\ddagger}$ & layerskip-llama2-70B(20)$^{\ddagger}$ \\

    \cline{2-6}
    
    & \multirow{2}{*}{Humaneval} & A40 & facebook & facebook & facebook \\
    & & $\times$2 & layerskip-codellama-34B & layerskip-codellama-34B(4)$^{\ddagger}$ & layerskip-codellama-34B(4)$^{\ddagger}$ \\
    
    \hline
    
    \multirow{10}{*}{Sec.\,\ref{subsec:robustness}} & \multirow{10}{*}{ShareGPT} & \multirow{10}{*}{\shortstack[c]{A100\\$\times$2}} & \multirow{10}{*}{\shortstack[c]{Qwen/ \\ Qwen2.5-32B-Instruct}} & Qwen/ & Qwen/ \\
    & & & & Qwen2.5-3B-Instruct & Qwen2.5-0.5B-Instruct \\ 
    \cline{5-6}
    & & & & \multirow{4}{*}{\shortstack[c]{Qwen/\\Qwen2.5-1.5B-Instruct}} & Qwen/ \\
    & & & &  & Qwen2.5-0.5B-Instruct \\ 
    \cline{6-6}
    & & & & & unsloth/ \\
    & & & & & Qwen2.5-0.5B-Instruct \\ 
    \cline{5-6}
    & & & & \multirow{4}{*}{\shortstack[c]{Qwen/\\Qwen2.5-0.5B-Instruct}} & Qwen/ \\
    & & & &  & Qwen2.5-1.5B-Instruct \\ 
    \cline{6-6}
    & & & & & unsloth/ \\
    & & & & & Qwen2.5-0.5B-Instruct \\
    \hline

\end{tabular}

\raggedleft\footnotesize{$^{\ddagger}$ Layerskip with (N) layers \phantom{0000}}
  \caption{Model Pair and Hardware Settings used in Evaluations.}
\end{table*}

\section{Evaluation Settings}

\subsection{Models and hyperparameters}
\label{appendix:models}
To demonstrate the versatility and broad applicability of SV, we selected two widely-used open-source LLM families: the Qwen and Llama series. Our experimental design encompasses three dimensions of variation per family: base models (no fine-tuning), instruction-tuned models, and task-tuned models.

\paragraph{Models used in overall evaluations}
\begin{itemize}
  \item \textbf{Large-sized models:} We employed the Qwen2.5-Instruct family\,\cite{yang2024qwen2} for large-sized evaluation in main experiments, we used the 72B variant as the \emph{target} model, the 1.5B variant as the \emph{companion} model, and the 0.5B variant as the \emph{draft} model.
  \item \textbf{Mid-sized models:} We selected the CodeLlama-34B \cite{roziere2023code} as target model for mid-scale experiments. and TinyLlama\_v1.1\_math\_code (1.2B) \cite{zhang2024tinyllama} as the \emph{companion}, and AMD-Llama-135M \cite{amd2024llama135m} as the \emph{draft}.
  \item \textbf{Small-sized models:} To cover smaller models, we included the Llama2-14B variant\,\cite{touvron2023llama2} as the \emph{target} model, paired with a JackFram\_llama-160m \cite{miao2024specinfer} as the \emph{companion}, and JackFram\_llama-68m \cite{miao2024specinfer} as the \emph{draft}, which was specifically trained for speculation tasks with a reduced parameter size.
\end{itemize}

\paragraph{Models used in SD variants evaluations}
\begin{itemize}
  \item \textbf{Large-sized models:} We adopted Llama-3.3-70B-Instruct\,\cite{grattafiori2024llama} as the \emph{target} model, paired with Qwen3-1.7B\,\cite{yang2025qwen3} as the \emph{companion} model, and Qwen3-0.6B\,\cite{yang2025qwen3} as the \emph{draft} model.
  \item \textbf{Mid-sized models:} We used Qwen3-32B\,\cite{yang2025qwen3} as the \emph{target} model, together with Llama3.2-3B-Instruct\,\cite{grattafiori2024llama} as the \emph{companion} model, and Llama3.2-1B-Instruct\,\cite{grattafiori2024llama} as the \emph{draft} model.
  \item \textbf{LayerSkip:} We also evaluated Layerskip-Llama2-70B and Layerskip-CodeLlama-34B to assess performance of SV adopted on self-speculation techniques. The number of drafting layers was set to the default values specified by model providers in huggingface repository\,\xxx{\cite{facebook2024layerskip, facebook2024layerskipcodellama34b}}.
  \item \textbf{Eagle-3:} We evaluated \textsc{Eagle-3}\,\cite{li2025eagle} by pairing each backbone (verifier) with its corresponding Eagle-3 speculator checkpoint, following the draft model checkpoint recommended in the official implementation\,\cite{safeailab2025eagle3impl, speculators2025}. When adopting SV on \textsc{Eagle-3}\,\cite{li2025eagle}, we used smaller companion models for speculation: Llama-3.2-1B-Instruct for the large-sized setting and Qwen3-0.6B for the mid-sized setting.
\end{itemize}

\begin{figure*}[!t]
  \centering
  \includegraphics[width=\textwidth]{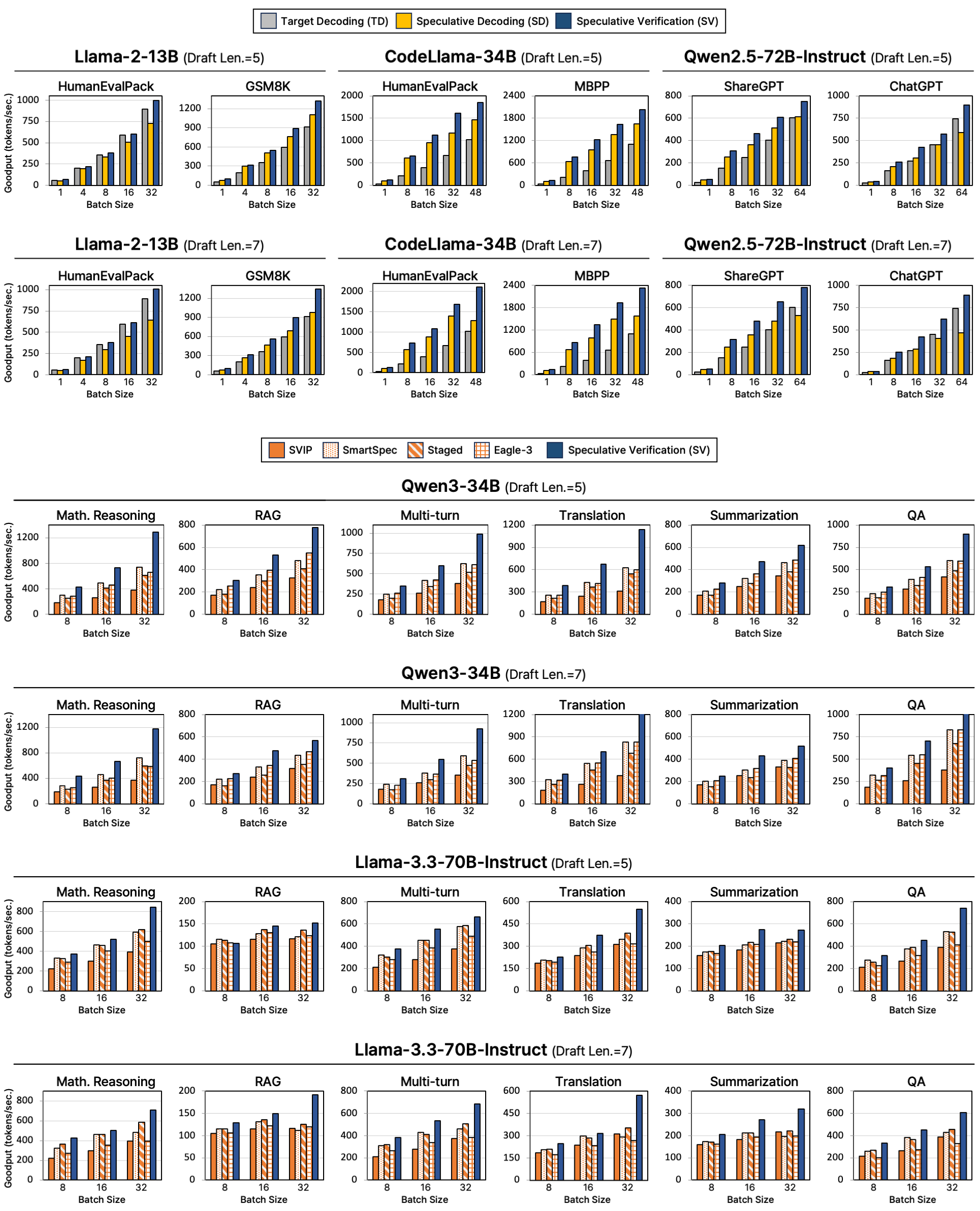}
  \caption{Performance Evaluation Across Model Pairs, Datasets, and Baselines.}
  \label{fig:appendix_performance}
\end{figure*}

\paragraph{Models used in other evaluations and analysis}
For additional analytical experiments (Section \ref{subsec:breakdown}, \ref{subsec:fairness}, and \ref{subsec:robustness}), we utilized the 32B/1.5B/0.5B model sizes of Qwen2.5-Instruct family. To test robustness across fine-tuned variants, we incorporated the fine-tuned version of Qwen2.5-Instruct models by unsloth\,\cite{unsloth2024qwen25}.

\paragraph{Sampling Parameters}
For all models, we adhered to the sampling hyperparameters recommended in their respective HuggingFace repositories or official GitHub documentation. Specific values are shown in Table\,\ref{tab:sampling_params}. \xxx{While many prior SD works report greedy decoding evaluations (or both greedy and sampling), we exclude greedy decoding because it often produces lower-quality outputs (e.g., repetitive tokens) and is less representative of typical LLM serving settings.}

\begin{table}[h]
  \centering
  \small
  \begin{tabular}{lcccc}
    \hline
    \multirow{2}{*}{\textbf{Model series}} & \multirow{2}{*}{\textbf{Top-K}}   & \multirow{2}{*}{\textbf{Top-P}} & \multirow{2}{*}{\textbf{Temp.}} & \textbf{Repetition} \\ 
    & & & & \textbf{Penelty} \\
    
    \hline
    Qwen2.5 based  & 20 & 0.8 & 0.7  & 1.05 \\
    Qwen3 based    & 20 & 0.6 & 0.95 & —    \\
    Llama-2 based  & —  & 0.9 & 0.6  & —    \\
    Llama-3 based  & —  & 0.9 & 0.6  & —    \\
    \hline
  \end{tabular}
  \caption{Sampling hyperparameters for each model series.}
  \label{tab:sampling_params}
\end{table}

\subsection{Dataset and Pre-Processing}

\begin{table}[t]
    \centering
    \small
    \setlength{\tabcolsep}{4pt}
    \renewcommand{\arraystretch}{1.15}
    \begin{tabular}{l p{0.68\columnwidth}}
    \hline
    \textbf{Dataset} & \textbf{Task} \\
    \hline
    GSM8K &
    Mathematical Reasoning \\
    
    HumanEvalPack &
    Code generation \\
    
    MBPP &
    Code generation \\
    
    ShareGPT &
    Multi-turn conversation \\
    
    ChatGPT &
    Multi-turn conversation \\
    
    Spec-Bench &
    Translation, Summarization,\newline
    QA, Mathematical Reasoning,\newline
    RAG, Multi-turn Conversation\\
    \hline
    \end{tabular}
    \caption{Datasets and corresponding tasks used in our evaluation.}
    \label{tab:datasets_tasks}
\end{table}

\begin{table*}[!t]
  \centering
  \setlength\tabcolsep{1pt}
  \small
  \renewcommand{\arraystretch}{1.1}
  \footnotesize

\begin{tabular}{ccll}
    \hline
    \textbf{Target Model} & \textbf{Symbol} & \textbf{Companion Model {\it (Training Entity)}} & \textbf{Information Gain (Draft Model}: Gain\textbf{)} \\
    \hline
    
    \multirow{6}{*}{\shortstack[c]{meta-llama/\\Llama-2-13b-chat-hf}} 
    & \textbf{A} & Llama-2-7b-hf ({\it Meta}) & \textbf{B}: 0.47, \textbf{C}: 0.52, \textbf{D}: 0.39, \textbf{E}: 0.57, \textbf{F}: 0.56 \\
    
    & \textbf{B} & TinyLlama\_v1.1 ({\it TinyLlama}) & \textbf{A}: 0.07, \textbf{C}: 0.39, \textbf{D}: 0.40, \textbf{E}: 0.33, \textbf{F}: 0.29 \\ 
    
    & \textbf{C} & Llama-160M-Chat-v1 ({\it Felladrin}) {\tiny finetuned from D} & \textbf{A}: 0.10, \textbf{B}: 0.13, \textbf{D}: 0.44, \textbf{E}: 0.40, \textbf{F}: 0.40 \\
    
    & \textbf{D} & llama-160m ({\it JackFram}) & \textbf{A}: 0.15, \textbf{B}: 0.05, \textbf{C}: 0.28, \textbf{E}: 0.59, \textbf{F}: 0.49 \\ 
    
    & \textbf{E} & llama-135m ({\it AMD}) & \textbf{A}: 0.11, \textbf{B}: 0.11, \textbf{C}: 0.32, \textbf{D}: 0.45, \textbf{F}: 0.42 \\
    
    & \textbf{F} & llama-68m ({\it JackFram}) & \textbf{A}: 0.11, \textbf{B}: 0.31, \textbf{C}: 0.40, \textbf{D}: 0.46, \textbf{E}: 0.37 \\
    
    \hline

    \multirow{6}{*}{\shortstack[c]{Qwen/\\Qwen2.5-14B-Instruct}} 
    & \textbf{A} & Qwen2.5-1.5B-Instruct ({\it Qwen}) & \textbf{B}: 0.36, \textbf{C}: 0.22, \textbf{D}: 0.18, \textbf{E}: 0.18, \textbf{F}: 0.22 \\

    & \textbf{B} & Qwen2.5-1.5B-Instruct ({\it Unsloth}) {\tiny finetuned from A} & \textbf{A}: 0.36, \textbf{C}: 0.22, \textbf{D}: 0.18, \textbf{E}: 0.18, \textbf{F}: 0.22 \\ 

    & \textbf{C} & TinySwallow-1.5B-Instruct ({\it SakanaAI}) {\tiny finetuned from A} & \textbf{A}: 0.12, \textbf{B}: 0.12, \textbf{D}: 0.12, \textbf{E}: 0.12, \textbf{F}: 0.21 \\

    & \textbf{D} & Qwen2.5-0.5B-Instruct ({\it Qwen}) & \textbf{A}: 0.03, \textbf{B}: 0.60, \textbf{C}: 0.06, \textbf{E}: 0.60, \textbf{F}: 0.54 \\ 

    & \textbf{E} & Qwen2.5-0.5B-Instruct ({\it Unsloth}) {\tiny finetuned from D}& \textbf{A}: 0.60, \textbf{B}: 0.03, \textbf{C}: 0.06, \textbf{D}: 0.03, \textbf{F}: 0.54 \\

    & \textbf{F} & Qwen2.5-0.5B-Instruct-ITA ({\it ReDiX}) {\tiny finetuned from D}& \textbf{A}: 0.33, \textbf{B}: 0.33, \textbf{C}: 0.11, \textbf{D}: 0.33, \textbf{E}: 0.33 \\

    \hline

    \multirow{6}{*}{\shortstack[c]{meta-llama/\\Llama-3.3-70B-Instruct}} 
    & \textbf{A} & Llama-3.2-3B-Instruct-pythonic ({\it Baseten}) & \textbf{B}: 0.23, \textbf{C}: 0.31, \textbf{D}: 0.09, \textbf{E}: 0.22, \textbf{F}: 0.25 \\
    
    & \textbf{B} & Llama-SmolTalk-3.2-1B-Instruct ({\it prithivMLmods}) & \textbf{A}: 0.04, \textbf{C}: 0.05, \textbf{D}: 0.04, \textbf{E}: 0.33, \textbf{F}: 0.39 \\ 
    
    & \textbf{C} & DeepSeek-R1-Distill-Llama-3B ({\it Suayptalha}) & \textbf{A}: 0.13, \textbf{B}: 0.17, \textbf{D}: 0.18, \textbf{E}: 0.13, \textbf{F}: 0.24 \\
    
    & \textbf{D} & Llama-3.2-1B-Instruct ({\it Unsloth}) & \textbf{A}: 0.05, \textbf{B}: 0.40, \textbf{C}: 0.04, \textbf{E}: 0.05, \textbf{F}: 0.35 \\ 
    
    & \textbf{E} & Llama-3.2-3B-Instruct ({\it Unsloth}) & \textbf{A}: 0.09, \textbf{B}: 0.23, \textbf{C}: 0.31, \textbf{D}: 0.22, \textbf{E}: 0.25 \\
    
    & \textbf{F} & Vikhr-Llama-3.2-1B-Instruct ({\it Vikhrmodels}) & \textbf{A}: 0.03, \textbf{B}: 0.30, \textbf{C}: 0.06, \textbf{D}: 0.03, \textbf{E}: 0.26 \\
    
    \hline
\end{tabular}

\caption{Uncertainty in token acceptance probability and information gain from observing $S$ and $A$. Symbols \textbf{A}--\textbf{F} denote the corresponding draft (or companion) models.}
\label{tab:info-gain-combinations}
\end{table*}

We evaluated our approach using seven distinct task categories: mathematical reasoning, code generation, multi-turn conversation, translation, summarization, QA, and RAGFor all experiments, we maintained consistency by using identical randomly sampled subsets across different evaluation scenarios.

For probability profile construction, we extracted 512 samples from training sets where available. For datasets without explicit train-test splits, we randomly sampled 512 instances. For goodput evaluation, we randomly selected between 128 and 256 samples from evaluation/test sets, carefully excluding any samples that appeared in the probability profile to prevent data leakage. These randomly sampled datasets remained constant across all experimental conditions to ensure fair comparisons.

For dialogue evaluation, we utilized two comprehensive datasets: ShareGPT\,\cite{sharegpt_vicuna_unfiltered}, a collection of human-assistant conversations extracted from various online sources, and ChatGPT Dataset\,\cite{chatgpt_prompts_dataset}, consisting of diverse dialogue prompts and responses.

Our code generation evaluation encompassed six programming languages using the HumanEvalPack \cite{muennighoff2023octopack} benchmark, which includes Python, C++, Java, JavaScript, Rust, and Go. We constructed a balanced subset by randomly sampling tasks across all languages to ensure comprehensive coverage of different programming paradigms and syntactic structures. We also incorporated MBPP \cite{austin2021program}, which consists of approximately 1,000 crowd-sourced Python programming problems. 

To assess mathematical reasoning capabilities, we employed the GSM8K\,\cite{cobbe2021gsm8k} dataset, which contains grade school math word problems that require multi-step reasoning to solve.

For Spec-Bench evaluation, we performed a task-wise split to support both probability profile construction and evaluation. Specifically, for each task category in Spec-Bench, we randomly sampled 10\% of the queries (8 queries per task) to construct the probability profile, and used the remaining 90\% for evaluation (72 queries per task). The same random partition was reused across all experimental settings to ensure consistency and fair comparison.

\subsection{Hardware Settings}
All experiments were conducted in multiple computing environments to meet the heterogeneous compute and memory demands of the evaluated models.

\paragraph{Azure cloud instances.}
For large-sized models, we used an Azure VM of type Standard NC96ads A100 v4. For mid- and small-sized models, we used Standard NC48ads A100 v4 instances. Both VMs are equipped with an AMD EPYC 7V13 64-core CPU and 2TB of RAM.

\paragraph{Private GPU server (LayerSkip experiments).}
LayerSkip experiments were performed on a private GPU server equipped with an AMD EPYC 7313 16-core CPU, four NVIDIA A40 GPUs (48\,GB VRAM each), and 500GB of RAM.

\section{Full Evaluation Results}

\subsection{Additional Evaluation Results for Section \ref{subsec:indicator} (Correlation of $S$)}
\label{appendix:correlation}

Table\,\ref{tab:correlation} reports the correlation values of $S$ for three representative draft/companion/target model combinations across ShareGPT, HumanEvalPack, and ChatGPT. Across all settings, the measured correlations are consistently high, ranging from 0.7249 to 0.8706. The strongest correlation is observed on ShareGPT with the Llama-based combination (0.8706), followed by HumanEvalPack with the code-specialized combination (0.8215). Even on ChatGPT, where the task and model setting differ substantially, the correlation remains strong at 0.7249. These results suggest that $S$ captures a stable relationship between the draft and companion models across different datasets and model families, supporting its use as a reliable indicator in SV.

\begin{table}[!h]
\centering
\small
    \renewcommand{\arraystretch}{1.1}
    \setlength\tabcolsep{4pt}
    
    \begin{tabular}{ccc}
    \hline
    \multirow{2}{*}{\textbf{Dataset}} & \textbf{Models} & \multirow{2}{*}{\textbf{Corr.}} \\
                                      & (D/C/T)        &   \\
    \hline
    
    \multirow{3}{*}{sharegpt}      &AMD-Llama-135m& \multirow{3}{*}{0.8706} \\
    & TinyLlama\_v1.1 & \\
    &  Llama-2-13b-hf& \\

    \hline
    \multirow{3}{*}{humanevalpack} & AMD-Llama-135m-code & \multirow{3}{*}{0.8215} \\
    & TinyLlama\_v1.1\_math\_code & \\
    &CodeLlama-34b-hf  & \\

    \hline
    \multirow{3}{*}{chatgpt} &Qwen2.5-0.5B-Instruct& \multirow{3}{*}{0.7249} \\
    & Qwen2.5-1.5B-Instruct & \\
    & Qwen2.5-72B-Instruct & \\
    \hline
    \end{tabular}
    
    \caption{Correlation results by dataset and model pairs (Target--Draft--Companion).}
    \label{tab:correlation}
\end{table}

\subsection{Additional Evaluation Results for Section \ref{subsec:eval-overall} (Overall Performance) and \ref{subsec:eval-sdvariants} (Comparison to SD Variants)}
\label{appendix:additional-eval}

Across both evaluations, SV consistently improves goodput over TD and standard SD. The gains typically grow with batch size, indicating that SV better amortizes verification overhead. Notably, as draft length increases, SD often incurs higher verification cost (more drafted tokens to validate per step), which can reduce goodput; in contrast, SV more effectively controls this overhead by reducing uncertainty in token acceptance and thereby preserving its advantage across draft length 5, 7.

\begin{table}[t]
  \centering
  \small
  \setlength\tabcolsep{4pt}
  \renewcommand{\arraystretch}{1.1}
  \begin{tabular}{cccccccc}
    \hline
    \multirow{2}{*}{\textbf{Cases}} & \textbf{Request} & \textbf{Avg. Veri.} & \multicolumn{5}{c}{\textbf{$\gamma$}} \\ \cline{4-8} 
     &  \textbf{ID} & \textbf{Length} & \textbf{1} & \textbf{2} & \textbf{3} & \textbf{4} & \textbf{5} \\ \hline
                   & 125 & 2.85  & 6 & 8 & 3 & 4 & 6 \\
     Bottom 5 reqs & 390 & 2.88 & 5 & 3 & 4 & 1 & 5 \\
     (lowest avg.  & 141 & 2.90 & 3 & 3 & 0 & 2 & 3 \\
     veri. length) & 121 & 3.00 & 1 & 1 & 1 & 1 & 1 \\
                   & 441 & 3.11 & 5 & 8 & 4 & 1 & 10 \\ \hline
     
                  & 274 & 4.96 & 0 & 0 & 0 & 2 & 43 \\
    Top 5 reqs    & 44 & 4.95 & 0 & 0 & 0 & 1 & 19 \\
    (highest avg. & 211 & 4.94 & 0 & 0 & 1 & 1 & 47 \\
    veri. length) & 419 & 4.91 & 0 & 1 & 0 & 0 & 43 \\
                  & 62 & 4.89 & 0 & 0 & 2 & 1 & 43 \\ \hline
  \end{tabular}
  \caption{Top/bottom 5 requests with highest/lowest average values of verification length $\gamma$ and its selected counts while scheduling}
  \label{tab:worst-case-requests}
\end{table}

\subsection{Additional Evaluation Results for Section \ref{subsec:eval-infogain} (Information Gain)}
\label{appendix:info-gain-subsection}
To examine whether SV consistently provides positive information gain across diverse draft--companion--target (D--C--T) configurations, we evaluated 90 distinct model triplets in Table\,\ref{tab:info-gain-combinations}. Specifically, we selected three target models and, for each target, paired it with six smaller models that can serve as either the draft or the companion, resulting in $3 \times (6 \times 5) = 90$ ordered (D, C, T) combinations. The selected triplets cover both regimes where the parameter gap between the draft and companion is small and where it is large.

For each triplet, we measured the target model's acceptance probabilities with and without SV, and quantified how our two variables ($S$ and $A$) reduce uncertainty in acceptance. We compute the uncertainty reduction as the information gain in predicting acceptance when conditioning on $(S, A)$, compared to the baseline without SV. Across all configurations, we consistently observed positive information gain, indicating that SV reliably decreases uncertainty about acceptance over a wide range of model combinations.

\subsection{Additional Evaluation Results for Section\,\ref{subsec:fairness} (Scheduling Fairness)}
\label{appendix:worst-case-requests}

The detailed experimental results for Section\,\ref{subsec:fairness} are shown in Table\,\ref{tab:worst-case-requests}.

\section{\rev{Information Gain vs.\ Distribution Similarity Correlation}}
\label{appendix:infogain-vs-correlation}

\rev{Table\,\ref{tab:infogain-vs-corr} compares the information gain achieved by SV with the Pearson correlation between draft--companion and draft--target distribution similarities, for a representative subset of model combinations. Even when the correlation is low (e.g., 0.26 for Llama2 1.1B/68M/13B), SV still achieves substantial positive information gain (0.31). This supports the claim in Section\,\ref{sec:sv} that SV requires only positive information gain -- a strictly weaker condition than high correlation -- to reduce speculation uncertainty.}

\begin{table}[h]
  \centering
  \small
  \setlength\tabcolsep{4pt}
  \renewcommand{\arraystretch}{1.1}
  \begin{tabular}{lcc}
    \hline
    \multirow{2}{*}{\textbf{Model (D/C/T)}} & \textbf{Info.} & \multirow{2}{*}{\textbf{Corr.}} \\
    & \textbf{Gain} & \\
    \hline
    Llama2 (1.1B / 68M / 13B)                  & 0.31 & 0.26 \\
    Llama3 (1B-Python / 3B-Russian / 70B)       & 0.25 & 0.76 \\
    Llama3 (1B / 3B-Python / 70B)               & 0.43 & 0.54 \\
    Llama3 (1B / 3B-Russian / 70B)              & 0.26 & 0.53 \\
    Llama3 (1B / 3B / 70B)                      & 0.22 & 0.73 \\
    Llama3 (3B-Python / 3B / 70B)               & 0.43 & 0.36 \\
    \hline
  \end{tabular}
  \caption{\rev{Information gain vs.\ Pearson correlation between draft--companion (D--C) and draft--target (D--T) distribution similarity. Corr. = correlation of D--C and D--T divergence values.}}
  \label{tab:infogain-vs-corr}
\end{table}

\section{Optimality of SV’s Verification Length Selection}
\label{sec:theory}

\begin{definition} Let $S$ be a random variable representing natural divergence of $P_d$ and $P_c$:
\vspace{-1pt}
\begin{equation}
\small
S = \sum_{j \in \text{vocab}} \min(P_d(t_j), P_c(t_j))
\vspace{-1pt}
\end{equation}
where $P_d$ and $P_c$ are token probability distributions of the draft and companion model, respectively.
\end{definition}

\begin{definition}
Let $A$ be a random variable for $t_d$'s acceptance probability in the companion model:
\vspace{-1pt}
\begin{equation}
\small
A = \min\left(1, \frac{P_c(t_d)}{P_d(t_d)}\right)
\vspace{-1pt}
\end{equation}
where $t_d$ is the token generated by the draft model, and $P_d$ and $P_c$ are defined as in A.1.
\end{definition}

\begin{definition}
Let $T_i$ be a random variable following the probability distribution of $i$'th draft token and $P(T_i)$ is its acceptance probability in the target model:
\vspace{-1pt}
\begin{itemize}
    \item $P(T_i=t)$ is the probability that a generated token $t$ is accepted by the target model and
    \vspace{-1pt}
    \item $P(T_i=t|S,A)$ is its conditional probability when $S$ and $A$ are given.
    \vspace{-1pt}
\end{itemize}
\end{definition}

\begin{definition}
\label{def:prob}
Let $N$ be a random variable for the number of tokens accepted by the target model. 

The probability for $N$ given $\gamma$ is calculated as:
\vspace{-1pt}
\begin{equation}
\small
P_{\gamma}(N) =
    \begin{cases}
    P(T_{N+1}\!\neq\!t_{N+1}) \prod_{i=1}^{N}P(T_i\!=\!t_i)  & \text{if } N < \gamma,\\
    \prod_{i=1}^{\gamma} P(T_i\!=\!t_i)                               & \text{if } N = \gamma
    \end{cases}
\end{equation}

The expected number of accepted tokens when verifying $\gamma$ tokens is calculated as:
\vspace{-1pt}
\begin{equation}
\small
E(N|\gamma) = \sum_{i=1}^{\gamma} i \cdot P_{\gamma}(N=i)
\vspace{-1pt}
\end{equation}
where $P_{\gamma}(N=i)$ is the probability of accepting exactly $i$ tokens when verifying $\gamma$ tokens.
\end{definition}

\begin{definition}
Let goodput($\gamma$) be the goodput (i.e., the number of accepted tokens per unit time) when verifying $\gamma$ number of tokens, which is calculated as:
\vspace{-1pt}
\begin{equation}
\small
\text{Goodput}(\gamma) = \frac{N}{\text{Latency}(\gamma)}
\end{equation}
where $N$ is the number of tokens accepted when verifying $\gamma$ tokens.

To estimate goodput, we use the profiled latency and expected number of accepted tokens as following:
\vspace{-1pt}
\begin{equation}
\small
\hat{\text{Goodput}(\gamma)} = \frac{E(N|\gamma)}{\text{Latency}_{\text{profiled}}(\gamma)}
\end{equation}
\end{definition}

\begin{assumption}
The draft and companion models are reasonably aligned with the target model.
\end{assumption}

\begin{definition}
We define $\hat{P}(T_i)$, i.e., an estimator of $P(T_i)$, to be $P(T_i|S,A)$.
\end{definition}

\begin{definition}
\label{def:observedprob}
We obtain $P(T_i|S,A)$ by observing sample data and grouping (i.e., binning) acceptance probability according to the values of $S$ and $A$ as following:
\begin{equation}
\small
\begin{aligned}
&P(T_i|S,A) = \\
&\begin{cases}
E(P(T_i|s_{0} \leq S < s_{1},\,a_{0} \leq A < a_{1}))\\
E(P(T_i|s_{1} \leq S < s_{2},\,a_{1} \leq A < a_{2}))\\
\dots
\end{cases}
\end{aligned}
\end{equation}
We can reduce the variance of the observed acceptance probabilities by controlling the bin sizes of $S$ and $A$. 
Reducing this variance subsequently decreases the uncertainty associated with the acceptance probabilities.
\end{definition}

\begin{assumption}
The observed probability in Definition \ref{def:observedprob} constitutes an unbiased estimator of $P(T_i)$, as long as the observed distributions of $S$, $A$, and $P$ coincide with those encountered at inference time.
\end{assumption}

\begin{assumption}
\label{assump:latency}
Let $Latency(\gamma)$ be the end-to-end wall-clock latency required to process $\gamma$ tokens. and its finite difference is
\begin{equation}
\small
\Delta Latency(\gamma)\;=\; Latency(\gamma+1)\;-\; Latency(\gamma), 
\quad \gamma\ge 1.
\end{equation}

We assume the following properties.
\begin{enumerate}
\small
  \item There exists a threshold $\gamma_{0}$ such that $\Delta Latency(\gamma)$ is a constant value for all $\gamma \geq \gamma_0$:
        \[
          \Delta Latency(\gamma)\;=\;\Delta Latency(\gamma_{0}) \enskip \text{for } \gamma \ge \gamma_{0}
        \]
  \item $\Delta Latency(\gamma)$ is an increasing function for $\gamma$ smaller than the threshold $\gamma_{0}$:
        \[
          \Delta Latency(\gamma+1)\;>\;\Delta Latency(\gamma) \enskip \text{for } 1 \leq \gamma < \gamma_{0}
        \]
\end{enumerate}
From the above two properties, $\Delta Latency(\gamma)$ is a non-decreasing function for all $\gamma > 0$:
\begin{equation}
\small
  \Delta Latency(\gamma+1)\;\ge\;\Delta Latency(\gamma),
  \enskip \text{for } \gamma > 0
\end{equation}
\end{assumption}

\begin{lemma}
\label{lemma:concave}
The expected number of accepted tokens $E(N|\gamma)$ is increasing and concave with respect to $\gamma$:
\vspace{-10pt}
{\small
\begin{align}
\frac{d}{d\gamma}E(N|\gamma) & \geq 0 \\
\frac{d^2}{d\gamma^2}E(N|\gamma) & \leq 0
\end{align}
}
\begin{proof}
The first and second derivatives are computed as discrete differences, represented as $\Delta$:
{\small
\begin{align}
\frac{d}{d\gamma}E(N|\gamma) &= \Delta^1 E(N|\gamma) = E(N|\gamma+1) - E(N|\gamma) \\
\frac{d^2}{d\gamma^2}E(N|\gamma) &= \Delta^2 E(N|\gamma) \notag \\
&= \Delta^1 E(N|\gamma+1) - \Delta^1 E(N|\gamma) \notag \\
&= E(N|\gamma+2) - 2E(N|\gamma+1) + E(N|\gamma).
\end{align}
}
\\
\textbf{Part 1: Proof of $\Delta^1 E(N|\gamma) \geq 0$.}
Increasing $\gamma$ can only increase the number of accepted tokens, thus $\Delta^1 E(N|\gamma) \geq 0$.
\\
\\
\textbf{Part 2: Proof of $\Delta^2 E(N|\gamma) \leq 0$.}
{\small
\begin{align}
&\Delta^1 E(N|\gamma) = E(N|\gamma+1) - E(N|\gamma) \\
&= \sum_{k=1}^{\gamma+1} {k P_{\gamma+1}(k)} - \sum_{k=1}^{\gamma} {k P_{\gamma}(k)} \\
&= \underbrace{\sum_{k=1}^{\gamma-1} {k (P_{\gamma+1}(k) - P_{\gamma}(k))}}_{\alpha} \\
& \quad + \underbrace{\gamma P_{\gamma+1}(\gamma) + (\gamma+1) P_{\gamma+1}(\gamma+1) - \gamma P_{\gamma}(\gamma)}_{\beta}
\end{align}
}
\vspace{-10pt}
{
\small
\begin{align}
&\Delta^2 E(N|\gamma) = \Delta^1 E(\gamma+1) - \Delta^1 E(N|\gamma)\\
&= \sum_{k=1}^{\gamma-1} k \underbrace{\left[ P_{\gamma+2}(k) - 2 P_{\gamma+1}(k) + P_{\gamma}(k) \right]}_{(a)} \\
&\quad + \gamma \underbrace{\left[ P_{\gamma+2}(\gamma) - 2 P_{\gamma+1}(\gamma) + P_{\gamma}(\gamma) \right]}_{(b_{1})} \\
&\quad + (\gamma+1) P_{\gamma+2}(\gamma+1) - 2(\gamma+1) P_{\gamma+1}(\gamma+1) \\
&\quad + (\gamma+2) P_{\gamma+2}(\gamma+2) \quad \text{: Term } (b_{2})
\end{align}
}
\vspace{-10pt}
\\

$\Delta^2 E(N|\gamma)$ is $\leq 0$ because $(a) = 0$ and $(b_{1})+(b_{2}) \leq 0$.

\begin{itemize}
    \item \textbf{Term $(a)=0$:} The probability $P(N=k)$ for $k < \gamma$ is independent of the total verification length if it's already $>\!k$.
    \item \textbf{Terms $(b_{1})+(b_{2}) \leq 0$:} These represent the diminishing returns in expected tokens as $\gamma$ increases.
\end{itemize}

\emph{Proof of terms (b) + (c) $\leq$ 0:} 
Applying Def. \ref{def:prob}, we express probabilities in terms of $P_{\gamma}(\gamma)$:
{
\small
\begin{equation}
\begin{aligned}
P_{\gamma+1}(\gamma) &= P_{\gamma}(\gamma) P(T_{\gamma+1} \neq t_{\gamma+1}) \\
P_{\gamma+1}(\gamma+1) &= P_{\gamma}(\gamma) P(T_{\gamma+1} = t_{\gamma+1}) \\
P_{\gamma+2}(\gamma) &= P_{\gamma}(\gamma) P(T_{\gamma+1} \neq t_{\gamma+1}) \\
P_{\gamma+2}(\gamma+1) &= P_{\gamma}(\gamma) P(T_{\gamma+1} = t_{\gamma+1}) P(T_{\gamma+2} \neq t_{\gamma+2}) \\
P_{\gamma+2}(\gamma+2) &= P_{\gamma}(\gamma) P(T_{\gamma+1} = t_{\gamma+1}) P(T_{\gamma+2} = t_{\gamma+2}).
\end{aligned}
\end{equation}
}
Then, substituting these:
{
\small
\begin{equation}
\begin{aligned}
(b_{0})+(b_{1}) =& -P_{\gamma}(\gamma) [ \gamma P(T_{\gamma+1} \neq t_{\gamma+1}) \\
&+ P(T_{\gamma+1} = t_{\gamma+1}) P(T_{\gamma+2} \neq t_{\gamma+2}) ] \leq 0
\end{aligned}
\end{equation}
}
Thus, $\Delta^2 E(N|\gamma) \leq 0$.
\end{proof}
\end{lemma}

\begin{lemma}
The Goodput function is concave with respect to $\gamma$:
\begin{equation}
\small
\Delta^{2}Goodput(\gamma) \leq 0 \text{ for all } \gamma
\end{equation}

\begin{proof}
{\small
\begin{equation}
\begin{aligned}
&\Delta\text{goodput}(\gamma) = \frac{E(N|\gamma+1)}{L(\gamma+1)} -\frac{E(N|\gamma)}{L(\gamma)} \\
&= \frac{L(\gamma)E(N|\gamma+1)-E(N|\gamma)L(\gamma+1)}{L(\gamma+1)L(\gamma)} \\
&= \frac{L(\gamma)\Delta E(N|\gamma) -E(N|\gamma)\Delta L(\gamma)}{L(\gamma+1)L(\gamma)}
\end{aligned}
\end{equation}
}

Let $\Psi(\gamma) = L(\gamma)\Delta E(N|\gamma) -E(N|\gamma)\Delta L(\gamma)$, then:
{\small
\begin{equation}
\begin{aligned}
\Delta\Psi(\gamma) &= \Psi(\gamma+1)-\Psi(\gamma) \\
&= \underbrace{L(\gamma)\Delta^2 E(N|\gamma)}_{(a)\leq 0} + \underbrace{(-E(N|\gamma)\Delta^2 L(\gamma))}_{(b)\le0} \\
&\quad + \underbrace{\Delta L(\gamma)\Delta E(N|\gamma+1) -\Delta L(\gamma+1)\Delta E(N|\gamma)}_{(c)\le0}
\end{aligned}
\end{equation}
}
All terms $(a), (b), (c)$ are $\leq 0$ based on Lemma \ref{lemma:concave} and Assumption \ref{assump:latency}. Thus $\Delta\Psi(\gamma) < 0$, meaning $Goodput(\gamma)$ is concave.

Optionally, with conditions:
{
\begin{align}
\Delta Goodput(\gamma_{0}) > 0, \quad \Delta Goodput(\gamma_{1}) < 0
\end{align}
}
we can find an optimal $\gamma^{*}$ that maximizes Goodput. Our evaluations indicate these conditions generally hold true.
\end{proof}
\end{lemma}

\end{document}